\documentclass{article}
\usepackage{geometry}
\usepackage[utf8]{inputenc} 
\usepackage[T1]{fontenc}    
\usepackage{hyperref}       
\usepackage{url}            
\usepackage{booktabs}       
\usepackage{amsfonts}       
\usepackage{nicefrac}       
\usepackage{microtype}      
\usepackage{xcolor}         
\usepackage{array}
\usepackage{amsmath}
\usepackage{amssymb}
\usepackage{graphics}
\usepackage{graphicx}
\usepackage{subfig}
\usepackage{xcolor}
\graphicspath{{figures_new/}}
\usepackage{amsthm}
\usepackage{algorithm2e}

\usepackage[round]{natbib}

\title{{Online Student-$t$ Processes with an Overall-local Scale Structure for Modelling Non-stationary Data}}

\author{Taole Sha \and Michael Minyi Zhang}

\begin{document}

\maketitle

\begin{abstract}
Time-dependent data often exhibit characteristics, such as non-stationarity and heavy-tailed errors, that would be inappropriate to model with the typical assumptions used in popular models. Thus, more flexible approaches are required to be able to accommodate such issues. To this end, we propose a Bayesian mixture of student-$t$ processes with an overall-local scale structure for the covariance. Moreover, we use a sequential Monte Carlo (SMC) sampler in order to perform online inference as data arrive in real-time. We demonstrate the superiority of our proposed approach compared to typical Gaussian process-based models on real-world data sets in order to prove the necessity of using mixtures of student-$t$ processes.
    
\end{abstract}

\section{Introduction}
In modelling dynamical systems, it is common that the data will exhibit non-stationarity, where the trend changes across the input space. Kernel methods like the Gaussian process (GP) are a popular choice of prior distribution over real-valued functions in Bayesian models of time series data \citep{GP}. However, in the non-stationary time series setting that this paper focuses on, they face several challenges: 1.) The calculation of the likelihood in GP inference requires inverting an $N\times N$ matrix, which generally incurs a computational complexity of $\mathcal{O}(N^3)$ where $N$ is the number of observations; 2.) Updating the model in real-time is not trivial; 3.) Stationarity is often assumed by covariance kernels, while non-stationary kernels typically lead to computationally intractable GPs, especially when the sample size is large. 
As a related stochastic process, the student-$t$ process (TP) has been treated as an attractive alternative prior distribution over function space compared to the GP \citep{t_as_alternative}. 
The TP is a more general elliptical process, where the likelihood of observations decreases in their distance from the mode, which is a reasonable assumption for the prior. It also has heavy tails controlled by the degree of freedom parameter, allowing more modelling flexibility. 

Similar to the GP, the TP has consistent marginals and closed-form conditionals which make it as convenient as the GP to use in statistical modelling without any additional computational cost. 
However, TPs are still liable to suffer from the aforementioned three issues that GPs face when modelling real-world data. \textit{Hence, we introduce in this paper a mixture of TPs with an SMC sampler, so that we may take advantage of the additional flexibility of a mixture-of-experts model with a convenient online inference algorithm.} 
To derive the TP, we assume a latent GP and integrate out an inverse gamma prior on the kernel amplitude and the noise parameters. Moreover, we add an additional parameter of the noise term to control the heteroscedasticity. 
Lastly, we model the level of heavy-tailedness by automatically controlling the TPs' degree of freedom using an efficient slice sampling scheme.

Our paper proceeds as follows: In Section \ref{sec:related_work}, we discuss some previous work about online GP models. We introduce our way of handling noisy data using TP in Section \ref{sec:student-t}. The online TP inference algorithm is detailed in Section \ref{sec:TP-MOE}. We use the experiment results to compare it with GP-based models in Section \ref{sec:experiments}. Finally, we conclude the paper in Section \ref{sec:conclusion} with a discussion of future work.

\section{Related Work}\label{sec:related_work}
The Gaussian process is a typical choice of prior used in Bayesian methods for modelling time series and non-linear regression problems \citep{GP}. A GP distributed function, $f \sim \mathcal{GP}(\mu(\cdot), \Sigma(\cdot,\cdot )$, is defined by a mean function, $\mu(\cdot)$, and a covariance function, $\Sigma(\cdot,\cdot)$, with a property that GPs are multivariate normally distributed conditioned on a finite set of points: $f(x) \sim \mathcal{N} (\mu(x), \Sigma(x, x^\prime)) $. While the GP is a convenient choice of prior due to this multivariate normal property, as it leads to tractable posterior inference in many classes of models, GPs suffer from the typical cubic computational that other kernel methods face. 

Numerous scalable methods have been developed to tackle the computational issue of GPs: Sparse inducing point methods are a popular technique for reducing the computational complexity of GP methods \citep{snelson2006sparse,titsias2009variational,bauer2016understanding}. In the sparse GP methods, they form a low-rank approximation of the kernel function using a collection of $M$ ``pseudo-inputs'' which reduce the computational complexity of the GP to $O(NM^2)$ from $O(N^3)$. Product-of-expert models employ a block diagonal approximation of the full covariance matrix in order to reduce the complexity of the full covariance matrix inversion to individually inverting each smaller block \citep{deisenroth2015distributed,cohen2020healing}. While not necessarily faster, mixture-of-expert models use a mixture of GPs to model functions with greater flexibility compared to a single GP \citep{rasmussen2001infinite,meeds2005alternative}.

For fast online GP methods, \cite{related_2002} used variational inference to approximate the posterior in a sparse online GP model, however, the hyperparameters are assumed to be fixed in their method. \cite{related_2008} proposed a product-of-experts local GP method for online fitting, where the weights are based on the distance of the new observation to the local models. Though, in such methods, ignoring the correlation between experts when adopting the local assumption can lead to poor uncertainty quantification. \cite{osvgp} developed a sparse variational GP regression approach that allows for online updating of the hyperparameters, called OSVGP. However, OSVGP has a tendency to be numerically unstable and, empirically, is liable to underfit the data. \cite{wiski} developed an exact sparse online model called WISKI, where a structured and sparse covariance matrix approximation developed by \cite{wiski's_kernel} is used, leading to constant computational complexity with respect to the number of observations. 

Regarding SMC methods in GPs, \cite{svensson2015marginalizing} proposed an SMC sampler with the purpose of marginalizing the kernel hyperparameters and \cite{gramacy2011particle} proposed an SMC sampler for sequential design in GPs. While these SMC methods allow for updating the GP model sequentially, they cannot account for non-stationarity in the data, nor are they able to limit the computational cost of the model as the complexity still scales $O(N^3)$.  However, \cite{ISMOE} proposed an importance sampling method for scaling up a mixture-of-experts GP model to an average complexity of $O(N^3 / K^2)$ for non-stationary data. Later, \cite{GPMOE} and \cite{harkonen2022mixtures} developed an online SMC and SMC$^2$ sampler for mixture of GPs. But despite the advances in online mixtures of GPs, little attention has been paid to online mixtures of the student-$t$ process.


\section{Student-$t$ Process for Noisy Data}\label{sec:student-t}
Student-$t$ priors have long been used in Bayesian linear regression problems, where we may desire modelling sparse regression coefficients or heavy-tailed errors \cite{fernandez1999multivariate,tipping2001sparse,west1984outlier,geweke1993bayesian}. \cite{vanhatalo2009gaussian} introduced a more robust method of GP regression where the latent function was GP distributed but the observation likelihood was assumed to be a student-$t$ distribution. However, they estimated all the parameters in this model using a Laplace approximation to the posterior distribution instead of performing exact Bayesian inference and therefore can properly capture the posterior uncertainty. Later, \cite{jylanki2011robust} used an expectation propagation algorithm for posterior inference in the same model. Again, expectation propagation is only an approximate method for posterior inference that cannot exactly capture the underlying uncertainty. 

When modelling the noisy time series data with $D$-dimensional input $\mathbf{x}_i\in \mathbb{R}^D$ and output $y_i\in \mathbb{R}$, where times $i=1,2,\ldots, N$, student-$t$ processes (TPs) are an attractive alternative to the Gaussian processes \citep{t_as_alternative}. Since student-$t$ distributions are not closed under addition, we cannot analytically obtain a latent TP and independent student-t noise for modelling regression problems.  As a result, this model formulation is not convenient for the statistical practitioner compared to the GP. 
Instead, we may obtain a TP by incorporating the noise into the kernel function. According to \cite{t_as_alternative}, \cite{t_process_wrong_example} wrongly assumed the noise to be independent when raising this model. \cite{t_as_alternative} stated that the noise term is uncorrelated but dependent, and argued it to behave similarly to a sum of a latent TP with independent noise. \cite{tang2017student} combined both a student-$t$ process model with student-$t$ noise, but, again, used only a Laplace approximation for the posterior instead of performing exact inference.


However, directly incorporating the noise term into the kernel may not be sufficiently flexible for modelling real data. In our model, we handle noisy data using an additional heteroscedastic parameter for each mixture. It is assumed that $y_i=f(\mathbf{x}_i)+\sigma_0\epsilon_i$, the output is generated by a latent zero-mean Gaussian Process $f(\mathbf{x}_i)$ and a Gaussian noise term where $\epsilon_i\sim N(0,\sigma_1^2)$. $\sigma_0^2$ is an overall scale parameter for both the covariance function $\sigma_0^2\Sigma(\cdot,\cdot)$ and the noise term, while $\sigma_1^2$ is a scale parameter for the noise term to control the heteroscedasticity. The GP and the noise are not independent here since they share the same overall scale. Any finite samples drawn from the GP at locations $\mathbf{X}=[\mathbf{x}_1,\cdots,\mathbf{x}_N]^T$ jointly follow a multivariate normal distribution:
\begin{align}
    f(\mathbf{X})|\mathbf{X},\sigma_0^2\sim \mathcal{N}_N(0,\sigma_0^2\Sigma_{\mathbf{X}\mathbf{X}^\prime}).
\end{align}
For noisy observations $\mathbf{y}=[y_1,\cdots,y_N]^T$, the data is generated by:
\begin{align}
    &\mathbf{y}|\mathbf{X},f,\sigma_0^2\sim \mathcal{N}_N(f(\mathbf{X}),\sigma_0^2\sigma_1^2\mathbf{I}).
\end{align}
Due to the conjugacy between Gaussians, marginally:
\begin{align}
\mathbf{y}|\mathbf{X},\sigma_0^2\sim\mathcal{N}_N(0,\sigma_0^2(\Sigma_{\mathbf{X}\mathbf{X}^\prime}+\sigma_1^2\mathbf{I})).
\end{align}
By marginalizing an inverse Gamma prior on $\sigma_0^2$ out, we can also arrive at the target multivariate student-t distribution with degree of freedom $\nu$:
\begin{align}
    &\sigma_0^2\sim \mbox{Inv\text{-}Gamma}\left(\frac{\nu}{2},\frac{\nu}{2}\right),\notag\\
    &\mathbf{y}|\mathbf{X}\sim \mathcal{T}_N(\nu, 0, \Sigma_{\mathbf{X}\mathbf{X}^\prime}+\sigma_1^2\mathbf{I}).
\end{align}

The TP's log marginal likelihood is:
\begin{align}
    &\log P(\mathbf{y}|\nu, \Sigma_{\mathbf{X}\mathbf{X}^\prime}, \sigma_1^2) = 
    -\frac{N}{2}\log(\nu\pi) \notag\\
    &~~~~~- \frac{1}{2} \log (|\Sigma_{\mathbf{X}\mathbf{X}^\prime} + \sigma_1^2\mathbf{I}|) + \log\left(\frac{\Gamma(\frac{\nu+N}{2})}{\Gamma(\frac{\nu}{2})}\right)\notag\\
    &~~~~~-\frac{\nu+N}{2}\log\left(1+\frac{\mathbf{y}^T(\Sigma_{\mathbf{X}\mathbf{X}^\prime} + \sigma_1^2\mathbf{I})^{-1}\mathbf{y}}{\nu}\right).
\end{align}
When making predictions, the posterior predictive distribution of $N^*$ target outputs $\mathbf{y}^*$ given new inputs $\mathbf{X}^*$ is:
\begin{align}
    &\Tilde{\phi}_2=\Sigma_{\mathbf{X}^{*}\mathbf{X}}(\Sigma_{\mathbf{X}\mathbf{X}^\prime}+\sigma_1^2\mathbf{I})^{-1}\mathbf{y},\notag\\
    &\beta_1=\mathbf{y}^T(\Sigma_{\mathbf{X}\mathbf{X}^\prime}+\sigma_1^2\mathbf{I})^{-1}\mathbf{y},\notag\\
    &\Tilde{\mathbf{K}}_{22}=(\Sigma_{\mathbf{X}^{*}\mathbf{X}^{{*}^\prime}}+\sigma_1^2\mathbf{I})\notag\\
    &~~~~~-\Sigma_{\mathbf{X}^{*}\mathbf{X}^\prime}(\Sigma_{\mathbf{X}\mathbf{X}^\prime}+\sigma_1^2\mathbf{I})^{-1}\Sigma_{\mathbf{X}\mathbf{X}^{*}},\notag\\
    &\mathbf{y}^{*}|\mathbf{y},\mathbf{X},\mathbf{X}^{*}\sim \mathcal{T}(\nu+N,\Tilde{\phi}_2,\frac{\nu+\beta_1}{\nu+N}\Tilde{\mathbf{K}}_{22}).
\end{align}

\section{Online Student-$t$ Processes for Non-stationary Data}\label{sec:TP-MOE}
The data generating process for our proposed model is:
\begin{align}
&\textbf{x}_i\sim \mathcal{T}(\boldsymbol{\mu}_{z_i},\boldsymbol{\Psi}_{z_i},\nu_{z_i}),\notag\\
&\alpha\sim \mbox{Gamma}(a_0,b_0),
~~z_i|\alpha\sim \mbox{CRP}(\alpha),\notag\\
&\boldsymbol{\theta}_k\sim log\mathcal{N}(m_0,s_0^2\mathbf{I}),
~~\nu_k \sim \mbox{Gamma}(2, 0.1),\notag\\
&h_k\sim \mathcal{N}(0,k_0^2),~~k_0^2\sim \mbox{Inv-Gamma}\left(\frac{1}{2}, \frac{1}{2}\right),\notag\\
&\mathbf{y}_k|\mathbf{X}_k,\boldsymbol{\theta}_k,\sim \mathcal{T}(\nu_k,0,\mathbf{K}_{\boldsymbol{\theta}_k}+|h_k|\mathbf{I}).
\end{align}
where the $i$-th input $\mathbf{x}_i$ comes from an infinite Dirichlet process Gaussian-inverse Wishart mixture model \citep{antoniak1974mixtures}: $\mathcal{N}(\mathbf{M}_{z_i}, \mathbf{C}_{z_i})$. The latent parameters $(\mathbf{M}_{z_i},\mathbf{C}_{z_i})$, for $z_i \in \{1, 2, \ldots \}$, are integrated out over a normal-inverse Wishart prior, $\mathcal{NIW}(\boldsymbol{\mu}_{z_i},\lambda_{z_i},\boldsymbol{\Psi}_{z_i},\nu_{z_i})$. The $x_i$ marginally follows a student-$t$ distribution $\mathcal{T}(\boldsymbol{\mu}_{z_i},\boldsymbol{\Psi}_{z_i},\nu_{z_i})$. 

The outputs $y_i$ from cluster $k$ are denoted as $\mathbf{y}_k$, which we obtain: 
\begin{align}
    &\sigma_k^2|\nu_k \sim \mbox{Inv-Gamma}\left(\frac{\nu_k}{2},\frac{\nu_k}{2}\right)\notag\\
    &\mathbf{y}_k|\sigma_k^{2},\mathbf{X}_k\sim \mathcal{N}(0,\sigma_k^2(\mathbf{K}_{\boldsymbol{\theta}_k}+|h_k|\mathbf{I})),\notag\\
    & \int P(\mathbf{y}_k|\sigma_k^{2},-)P(\sigma_k^{2}) d \sigma_k^{2} \sim T(\nu_k,0,\mathbf{K}_{\boldsymbol{\theta}_k}+|h_k|\mathbf{I}).
\end{align}

It is assumed to be a sum of a GP and dependent Gaussian noise as mentioned in Section \ref{sec:student-t}. The covariance is determined by an overall scale parameter $\sigma_k^2$ for both the kernel and the noise, a local scale parameter $h_k$ for the noise only to control this heteroscedasticity, and kernel parameters $\boldsymbol{\theta}_k$. The overall scale parameter $\sigma_k^2$ is integrated out over an inverse gamma prior, and a TP can be derived. When the $i^{th}$ streaming data $(\mathbf{x}_i,y_i)$ comes, we assign it to cluster $k$ according to the predictive distribution of the DP, the Chinese restaurant process \citep{aldous1985exchangeability}:
\begin{align}
P(z_{i} = k| \alpha, \mathbf {X}_{k}) \propto \left\lbrace \begin{array}{ll}N_{k}^{\prime } \cdot \mathcal {T}(\boldsymbol{\mu}_{k}^{\prime }, \boldsymbol{\Psi }_{k}^{\prime }, \nu _{k}^{\prime }) & k \in K^{+}. \\ \alpha \cdot \mathcal {T}(\boldsymbol{\mu}_{0}, \boldsymbol{\Psi }_{0}, \nu _{0}) & \text{o.w.} \end{array}\right.
\end{align}
$K^+$ refers to the existing clusters, and all $(\cdot)^\prime$ represent summary statistics calculated with first $i-1$ observations. The student-t likelihood's parameters $(\mathbf{\mu}_k^{'},\mathbf{\Psi}_k^{'},\nu_{k}^{'})$ of inputs can be updated by:
\begin{align} 
\boldsymbol{\mu}_{k}^{\prime } =& \frac{\lambda _{0}\boldsymbol{\mu}_{0} + N_{k}^{\prime } \bar{\mathbf {x}}_{k}}{\lambda _{k}^{\prime }}, \bar{\mathbf {x}}_{k}^{\prime } = \frac{\sum _{i^{\prime } : (z_{i^{\prime }} = k, i^{\prime } < i)} \mathbf {x}_{i^{\prime }} }{N_{k}^{\prime }},\notag\\ 
N_{k}^{\prime } =& \sum _{i^{\prime }=1}^{i-1}I(z_{i^{\prime }}=k), \lambda _{k}^{\prime } = \lambda _{0} + N_{k}^{\prime },\notag\\ 
\nu _{k}^{\prime } =& \nu _{0} + N_{k}^{\prime } - D + 1,\notag\\ 
\boldsymbol{\Psi }_{k}^{\prime } =& \frac{\lambda _{k}^{\prime }+1}{\lambda _{k}^{\prime }\nu _{k}^{\prime }} \left(\boldsymbol{\Psi }_{0} + \mathbf {S}_{k}^{\prime } + \mathbf {S}_{\bar{\mathbf {x}}_{k}}^{\prime } \right)\notag\\ 
\mathbf {S}_{k}^{\prime } =& \sum _{i^{\prime } : (z_{i^{\prime }} = k, i^{\prime } < i)} \left(\mathbf {x}_{i^{\prime }} - \bar{\mathbf {x}}^{\prime }_{k} \right)\left(\mathbf {x}_{i^{\prime }} - \bar{\mathbf {x}}^{\prime }_{k} \right)^{T}\notag\\ 
\mathbf {S}_{\bar{\mathbf {x}}_{k}}^{\prime } =& \frac{\lambda _{0} N_{k}^{\prime }}{\lambda _{k}^{\prime }} \left(\bar{\mathbf {x}}^{\prime }_{k} - \boldsymbol{\mu}_{0}\right)\left(\bar{\mathbf {x}}^{\prime }_{k} - \boldsymbol{\mu}_{0}\right)^{T}.  
\end{align}

Also, a Gamma prior is placed on the Dirichlet process concentration parameter $\alpha$. We can use a variable augmentation scheme to sample its full conditional posterior up to observation $i$ \citep{alpha_sampling}. 
\begin{align} 
\rho | \alpha \sim &\text{Beta}(\alpha +1, i), K = | \lbrace k : N_{k} > 0 \rbrace |\notag\\ 
\frac{\pi _\alpha }{1-\pi _\alpha } =&\frac{a_{0} + K -1}{N (b_{0} - \log \rho) }\notag\\ 
\alpha | \mathbf {z}_{1:i}, \pi _\alpha, \rho =&(1-\pi _\alpha)\notag\\
\cdot & \text{Gamma}(\alpha _{0} + K -1, b_{0} -\log \rho)\notag\\ 
+ & \pi _\alpha \cdot \text{Gamma}(\alpha _{0} + K, b_{0} -\log \rho). 
\end{align}

The $\mbox{Gamma}(2, 0.1)$ prior is commonly used when inferring the degree of freedom, which puts mass on a large range of reasonable values for the degrees of freedom \citep{gamma_prior_over_dof}. 
We sample the degrees of freedom parameter through an efficient variable augmentation scheme. 
Given the latent overall scale $\sigma_k^2$, the degree of freedom $\nu_k$ will be independent of all other parameters and data.
Due to the conjugacy between the Gaussian likelihood and the inverse Gamma prior, we can directly Gibbs sample the $\sigma_k^2$ from its full conditional. Then, conditioned on $\sigma^2_k$, we sample $\nu_k$ using the slice sampler from $P(\nu_k|\sigma_k^2)$ \citep{neal2003slice,damlen1999gibbs}. 
\begin{align}
    &\sigma_k^2|\mathbf{X}_k,\mathbf{y}_k,\nu_k \sim \mbox{Inv-Gamma}(\alpha^{\prime}_{\sigma^2}, \beta^{\prime}_{\sigma^2}),\notag\\
    & \alpha^{\prime}_{\sigma^2} = \frac{\nu_k+N_k^\prime}{2}\notag\\
    & \beta^{\prime}_{\sigma^2} = \frac{\nu_k+\mathbf{y}_k^T (\mathbf{K}_{\boldsymbol{\theta}_k}+|h_k|\mathbf{I})^{-1} \mathbf{y}_k}{2}
\end{align}
We assume a hierarchical structure on the local heteroscedasticity parameter, $|h_k|$, where global scale $k_0^2$ is shared over all mixtures.
Here, we will share scale data from other clusters to inform the posterior sampling of $h_k$. 
Because $h_k$ has a normal prior, we can again sample the full conditional of $k_0^2$ in closed form:
\begin{align}
    &k_0^2|h_1, \ldots, h_K \sim \mbox{Inv-Gamma}\left(\frac{K+1}{2}, \frac{1+\sum_{i=1}^{K}h_i^2}{2K}\right).
\end{align}
Then we sample $h_k$ and the TP parameters $\boldsymbol{\theta}_k$ using the elliptical slice sampler (ESS), which is an efficient sampling algorithm for non-conjugate models with Gaussian priors \citep{ellipticalss}. 



\subsection{SMC for Online TP-MOE}
In our proposed method, we use a sequential Monte Carlo sampler in order to update the model as new data arrive \citep{del2006sequential}. SMC follows from importance sampling (IS) and sequential importance sampling (SIS) algorithms in Monte Carlo methods, where IS and SIS sample the parameter of interest from a proposal distribution in order to approximate an intractable distribution:
\begin{align}
     \int P(X|\theta)P(\theta) \mbox{d}\theta \approx \frac{1}{J}\sum_{j=1}^{J}w^{(j)}\delta_{\theta^{(j)}}.
\end{align}
However, IS and SIS suffer from the particle degeneracy problem where one proposal weight, $w^{(j)}$, dominates the rest of the proposals. In SMC methods, we resample the particles with probability equal to the proposal weight. In this way, we replenish the sampler with particles that have high weight and remove particles that have low weight. 

For $j=1,\ldots,J$ particles, the particles $(\mathbf{z}^{(j)},\boldsymbol{\theta}^{(j)},\mathbf{h}^{(j)},\alpha^{(j)})$ are updated as described before when a new observation arrives. Then, we calculate the particle weights, which results in a posterior weighted sample TP product-of-experts models. Initially when $i=1$, the particle $j$'s weight is: 
\begin{align}
    w_{1}^{(j)} \propto P(y_{1} | z_{1}^{(j)}, \mathbf {x}_{1}, \boldsymbol{\theta }^{(j)}, h^{(j)}, \nu^{(j)}) P(\mathbf {x}_{1} | z_{1}^{(j)}, \alpha ^{(j)}). 
\end{align}
Then the updating procedure for $i>1$ is shown in Algorithm \ref{alg:SMC}.

\RestyleAlgo{ruled}
\SetKwComment{Comment}{/* }{ */}
\begin{algorithm}[hbt!]
    \caption{SMC Sampler for TP-MOE}
    \label{alg:SMC}
    \KwIn{New observation $(\mathbf{x}_i,y_i)$}
    \For{$j=1,\cdots,J$ in parallel}{
    Sample $z_i^{(j)} = k$ from $P(z_i^{(j)}|\alpha^{(j)},\mathbf{X}_{1:i-1})$\\
    Sample $\alpha^{(j)}$ from the full conditional $P(\alpha^{(j)}|\mathbf{z}_{1:i})$\\
    Sample $\theta_{k}^{(j)}$ and $h_{k}^{(j)}$ jointly by using the elliptical slice sampler\\
    Sample $(k_0^2)^{(j)}$ from $P((k_0^2)^{(j)}|h_1^{(j)}, \ldots, h_K^{(j)})$\\
    Sample $(\sigma_k^2)^{(j)}$ from $P(\sigma_k^2|\mathbf{X}_k,\mathbf{y}_k,\nu_{k}^{(j)})$\\
    Sample $\nu_{k}^{(j)}$ by using the slice sampler\\
    Update particle weight:
    \begin{align}
        w_i^{(j)}&= w_{i-1}^{(j)}P(\mathbf{x}_{i}|\alpha^{(j)},z_{i}^{(j)})\notag \\
        &\times \frac{P\left(\mathbf{y}_{1:i}|\mathbf{X}_{1:i},\mathbf{\theta}_{k,i}^{(j)},h_{k}^{(j)}, \nu_{k}^{(j)}\right)}{P\left(\mathbf{y}_{1:i-1}|\mathbf{X}_{1:i-1},\mathbf{\theta}_{k}^{\prime(j)},h_{k}^{\prime(j)}, \nu_{k}^{\prime(j)}\right)}
    \end{align}
    }
    Normalize weights:
    \begin{gather*}
        w_i^{(j)}:=\frac{w_i^{(j)}}{\sum_{j=1}^{J}{w_i^{(j)}}}
    \end{gather*}
    \If{$N_{eff}<\frac{J}{2}$}{
    Resample particles $(\mathbf{z}_{1:i}^{(\mathbf{j}^*)},\mathbf{\theta}_{k}^{(\mathbf{j}^*)},h_{k}^{(\mathbf{j}^*)},\nu_{k}^{(\mathbf{j}^*)},\alpha^{(\mathbf{j}^*)})$, where $\mathbf{j}^*\sim \mbox{Multinomial}(J,w_i^{(1)},\cdots,w_i^{(j)})$\\
    Set $w_i^{(j)}:=\frac{1}{J}\mbox{ for }j=1,\cdots,J$\\
    }
    \KwOut{Particle weights $(w_i^{(1)},\cdots,w_i^{(j)})$ and particles $(\mathbf{z}_{1:i}^{(1:J)},\mathbf{\theta}^{(1:J)},\mathbf{h}^{(1:J)},\mathbf{\nu}^{(1:J)},\alpha^{(1:J)})$}
\end{algorithm}
\begin{algorithm}[hbt!]
    \caption{TP-MOE Prediction}
    \label{alg:predict}
    \For{$j=1,\cdots,J$}{
    Predict new observations on particle j with
    \begin{align}
        &p_k\propto N_k\cdot P(\mathbf{x}_*|z_*=k,\mathbf{X}_k,-)\notag\\
        &P(y_*^{(j)}|\mathbf{y},\mathbf{X},\mathbf{x}_*,-)=\notag\\
        &~~~~~\sum_{k\in \mathbf{K}^+}{p_k\cdot P(y_{k*}^{(j)}|\mathbf{y}_k,\mathbf{X}_k,\mathbf{x}_*,-)}
    \end{align}
    }
    Average predictions: $P(\Bar{y}_*|\mathbf{y},\mathbf{X},\mathbf{x}_*)=\sum_{j=1}^J{w_i^{(j)}P(y_*^{(j)}|\mathbf{y},\mathbf{X},-)}$
\end{algorithm}

The computational complexity is dominated by the inversion of a $N_k \times N_k$ matrix. 
If we assume that the average size of $N_k$ is $N/K$, the number of data divided by the number of clusters, the n the computational complexity will be $\mathcal{O}(JN^3/K^2)$. Under the basic setting of our sampler, the complexity of the sampler still grows as new data arrive so the method cannot truly be considered ``online''. To this end, we adopt the ``minibatched'' stochastic approximation that is widely used as a method for substantially reducing the computational complexity of posterior inference \citep{GPMOE,ISMOE,minibatch_exa_1,minibatch_exa_2}. A subsample of size $B$ from the mixture with $N_k$ observations is drawn uniformly without replacement, then their likelihood is calculated and upweighted by $N_k/B$ power to approximate the full likelihood. The stochastic approximation method leads us to:
\begin{align} 
& \mathbf {u}_{k}^{(j)} \!=\! (u_{1}, \ldots, u_{B})\notag\\ 
&~~~~~\sim \text{HyperGeometric} \big(B, \big\lbrace i : z_{i}^{(j)}=k \big\rbrace\big), \notag\\ 
&(\mathbf {y}_{\mathbf {u}_{k}}, \mathbf {X}_{\mathbf {u}_{k}}) = \big(y_{u}, \mathbf {x}_{u} : u \in \mathbf {u}_{k}^{(j)}\big). \notag\\ 
& P \big(\mathbf {y}_{\mathbf {u}_{k}^{(j)}}| \mathbf {X}_{\mathbf {u}_{k}^{(j)}}, \boldsymbol{\theta }_{k}^{(j)}, h_k, \sigma_k^2 \big) \notag\\
&~~~~~\sim \mathcal {N} \left(0, \sigma_k^2 (\mathbf{K}_{\boldsymbol{\theta}_k^{(j)}} + \frac{N_{k}|h_{k}|}{B} \mathbf {I}) \right), \notag\\
& P \big(\mathbf {y}_{\mathbf {u}_{k}^{(j)}}| \mathbf {X}_{\mathbf {u}_{k}^{(j)}}, \boldsymbol{\theta }_{k}^{(j)}, h_k, \nu_k \big)  \notag\\
&~~~~~\sim \mathcal {T} \left(\nu, 0, \mathbf{K}_{\boldsymbol{\theta}_k^{(j)}} + \frac{N_{k}|h_{k}|}{B} \mathbf {I}\right). 
\end{align}
With minibatching, the complexity is reduced to $\mathcal{O}(J\text{min}\{N_k,B\}^3/K^2)$. As each particle can be updated independently, the parallel computation can be adopted to further reduce the complexity to $\mathcal{O}(\text{min}\{N_k,B\}^3/K^2)$. Then, we calculate the effective sample size, $N_{eff}=1/\sum_{j=1}^J(w_i^{(j)})^2$, based on the particle weights. If it is lower than a threshold set by the user, typically $J/2$, the particles are resampled to only preserve the high-weighted ones. We make predictions using a weighted average detailed in Algorithm \ref{alg:predict}.


\section{Experiments}\label{sec:experiments}
The choice of hyperparameters could significantly influence the TP-MOE's performance. According to \cite{ISMOE}, larger $J$ and $B$ will lead to better performance but more computation time, while in contrast, increasing $K$ will decrease both model performance and computation time. 
In this section, we proceed to study the advantages of the heavy tails by implementing the TP-MOE and other Gaussian-based models on different 
non-stationary datasets and analysing their performances in terms of one-step-ahead predictions. The GP models include a Gaussian mixture-of-experts model (GP-MOE) \citep{GPMOE},  a sparse online GP method using the Woodbury identity and structured kernel interpolation (WISKI) \citep{wiski}, and an online sparse variational GP method (OSVGP) \citep{osvgp}
\footnote{ The implementation for GP-MOE is available at \texttt{https://github.com/michaelzhang01/GPMOE}. The code of OSVGP and WISKI are available at: \texttt{https://github.com/wjmaddox/online\_gp}. Our code will be submitted in the supplementary material.}. 

For the experiments, we sequentially predict the next future observation and update the model with the real data point. The one-step predictive mean squared error (MSE) is adopted to evaluate the results. 
The 5 datasets used include: 1.) An accelerometer measurement of a motorcycle crash (N=94). 2.) The price of Brent crude oil (N=100). 3.) The annual carbon dioxide output in Canada (N=215). 4.) The annual water level of the Nile River data (N=100). 5.) The exchange rate between the Euro and the US Dollar (N=200) 
\footnote{The motorcycle dataset can be found in the R package \texttt{VarReg}. The Brent, Canada CO$_2$, and Nile River datasets are available at: \texttt{https://github.com/alanturing-institute/TCPD}. The EUR-USD dataset is available in the R package \texttt{priceR}.}. 
The first three exhibit non-stationarity in both length-scale and noise, while the Nile River dataset shows only time-varying mean values and the exchange rate dataset is a series of non-stationary noise. They have been pre-processed to have zero mean and unit variance.


To make the results comparable, The TP-MOE and the GP-MOE share the same particle number $J=100$ and the same $16$ cores used on a shared memory process based on OpenMP, and the number of inducing points for all models is set to be $50$. The OSVGP's number of optimization iterations is set to the default value of $1$. The radial basis function kernel for all models is:
\begin{align}
    \Sigma(\mathbf{x},\mathbf{x}^\prime)=\exp  \left\{-\frac{\theta}{2}\sum_{d=1}^D(x_d-x_d^\prime)^2\right\}
\end{align}

\begin{figure*}[htb!]
     \centering
     \vspace{.3in}
     \subfloat[TP-MOE]{\includegraphics[width = 0.25\textwidth]{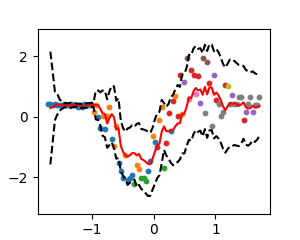}}
     \subfloat[GP-MOE]{\includegraphics[width = 0.25\textwidth]{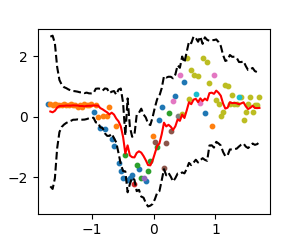}}
     \subfloat[WISKI]{\includegraphics[width = 0.25\textwidth]{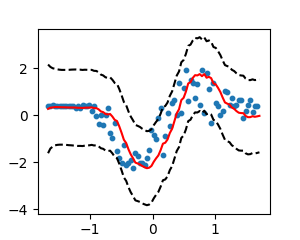}}
     \subfloat[OSVGP]{\includegraphics[width = 0.25\textwidth]{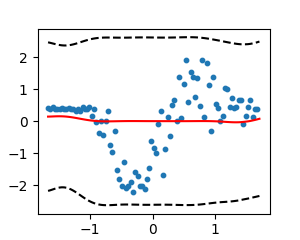}}
     \vspace{.3in}
     \caption{
     Sample Runs on the Motorcycle Dataset. N = 94.}
     \label{motor}
\end{figure*}

\begin{figure*}[htb!]
     \centering
     \vspace{.3in}
     \subfloat[TP-MOE]{\includegraphics[width = 0.25\textwidth]{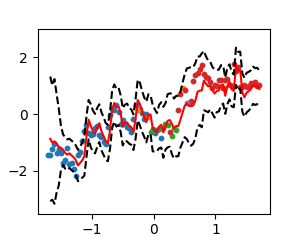}}
     \subfloat[GP-MOE]{\includegraphics[width = 0.25\textwidth]{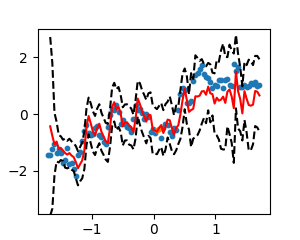}}
     \subfloat[WISKI]{\includegraphics[width = 0.25\textwidth]{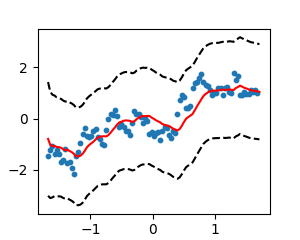}}
     \subfloat[OSVGP]{\includegraphics[width = 0.25\textwidth]{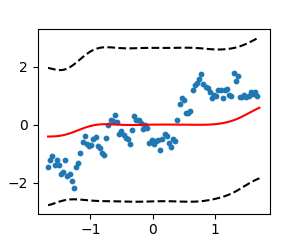}}
     \vspace{.3in}
     \caption{
     Sample Runs on the Brent Dataset. N = 100.}
     \label{brent}
\end{figure*}

\begin{figure*}[htb!]
     \centering
     \vspace{.3in}
     \subfloat[TP-MOE]{\includegraphics[width = 0.25\textwidth]{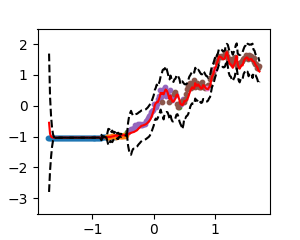}}
     \subfloat[GP-MOE]{\includegraphics[width = 0.25\textwidth]{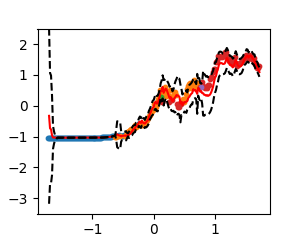}}
     \subfloat[WISKI]{\includegraphics[width = 0.25\textwidth]{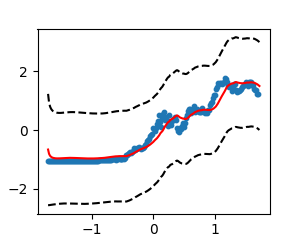}}
     \subfloat[OSVGP]{\includegraphics[width = 0.25\textwidth]{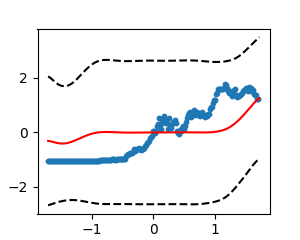}}
     \vspace{.3in}
     \caption{
     Sample Runs on the Canada Dataset. N = 215.}
     \label{canada}
\end{figure*}

\begin{figure*}[htb!]
     \centering
     \vspace{.3in}
     \subfloat[TP-MOE]{\includegraphics[width = 0.25\textwidth]{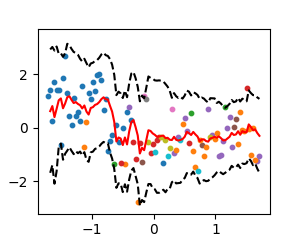}}
     \subfloat[GP-MOE]{\includegraphics[width = 0.25\textwidth]{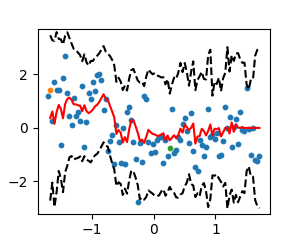}}
     \subfloat[WISKI]{\includegraphics[width = 0.25\textwidth]{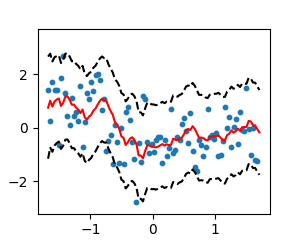}}
     \subfloat[OSVGP]{\includegraphics[width = 0.25\textwidth]{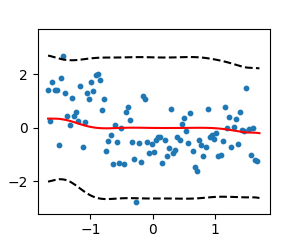}}
     \vspace{.3in}
     \caption{
     Sample Runs on the Nile River Dataset. N = 100.}
     \label{nile}
\end{figure*}

\begin{figure*}[htb!]
     \centering
     \vspace{.3in}
     \subfloat[TP-MOE]{\includegraphics[width = 0.25\textwidth]{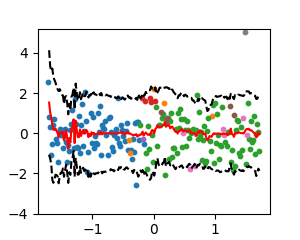}}
     \subfloat[GP-MOE]{\includegraphics[width = 0.25\textwidth]{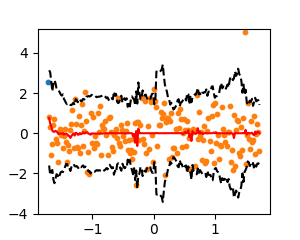}}
     \subfloat[WISKI]{\includegraphics[width = 0.25\textwidth]{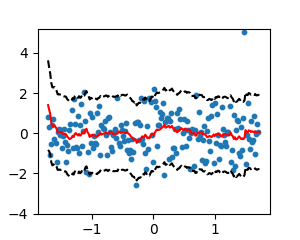}}
     \subfloat[OSVGP]{\includegraphics[width = 0.25\textwidth]{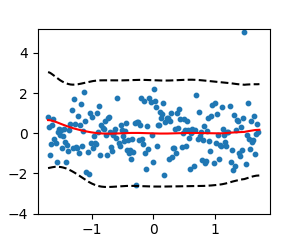}}
     \vspace{.3in}
     \caption{
     Sample Runs on the EUR-USD Dataset. N = 200.}
     \label{exchange}
\end{figure*}

The plots of four algorithms' sample runs are shown in Figure \ref{motor}-\ref{exchange}, which contain data points, one-step predictive mean (plotted with solid red lines) and 95\% predictive interval (plotted with dashed black lines). The data points in TP-MOE's and GP-MOE's plots are coloured according to the cluster assignment given by the particle with the highest weight. The results in terms of the predictive MSE are listed in Table \ref{tab:mse}.
\begin{table*}[htb!]
    \centering
\caption{One-step Predictive MSE. One Standard Error Reported in Parentheses}
\label{tab:mse}
    \begin{tabular}{llllll}
              &Motorcycle & Brent &Canada & Nile &EUR-USD\\
         \hline 
         TP-MOE& \textbf{0.363 (0.028)} &\textbf{0.146 (0.014)}& \textbf{0.015 (0.003)} & \textbf{0.738 (0.017)} & 1.028 (0.013)\\
         GP-MOE& 0.381 (0.038) &0.160 (0.019) & 0.016 (0.004) & 0.752 (0.025) & \textbf{1.004 (0.009)}\\
         WISKI& 0.631 (0.000) &0.220 (0.000) & 0.048 (0.000) & 0.767 (0.000) & 1.061 (0.000)\\
         OSVGP& 0.998 (0.002) &0.782 (0.021) & 0.711 (0.030) & 0.908 (0.008) & 1.019 (0.003)\\
    \end{tabular}
\end{table*}
From the comparisons in Table \ref{tab:mse}, we observe that our TP-MOE performs better than the GP-based models and achieves lower predictive MSE. 
According to the plots of sample runs (Figure \ref{motor}-\ref{canada}), 
we can see that the MOE models can better capture the heterogeneity of the underlying function better than the stationary models. Moreover, we can see that the TP-MOE produces tighter predictive credible intervals compared to the GP-MOE, which sometimes produces overly conservative predictive intervals which suggests that the TP-MOE has better uncertainty quantification capabilities.
The OSVGP tends to underfit as expected, while the WISKI cannot quantify the uncertainty as well as the mixture-of-experts models due to its assumption of stationarity.

For the Nile River data which exhibits only non-stationary mean values, the zero-mean assumption makes it hard for the TP-MOE and the GP-MOE to model the trend. Despite the model misspecification, the TP-MOE still achieves the best predictive MSE among four models according to Table \ref{tab:mse}. Also, based on the comparison in Figure \ref{nile}, the TP-MOE's 95\% predictive intervals are the most consistent with the trend. The heavy-tailed property helps it be more robust to the model misspecification.

However, when modelling the time-varying noise in the EUR-USD dataset, the GP-MOE handles this task the best (Table \ref{tab:mse}). Figure \ref{exchange} reveals the potential reason for the failure of TP-MOE here. the predictive means are not very stable, and tend to capture some trends that may not really exist. The OSVGP which performs poorly in previous datasets even achieves better results this time, because it usually underfits the streaming data and maintains stable predictions. 
Hence we conclude that, compared with the GP-based models, our TP-MOE can better fit the data that exhibit non-stationarity in length scale and noise, providing more accurate predictive means and most of the time better uncertainty quantification. Also, the heavy tails help it be more robust to the model misspecification. However, if the noise dominates the series, the kernelized covariance is less useful to capture such a trend.



\section{Conclision}\label{sec:conclusion}
Heavy-tailed data sets appear in a wide variety of applied settings. However, devising models that can adequately handle their noise structure is not trivial. In this paper, we build a Bayesian mixture of student-$t$ processes model with an overall-local scale structure for noisy data, which can be inferred by an SMC online algorithm. We have shown that TP-MOE has advantages over the Gassian-based models when facing commonly encountered non-stationary data.

In future work, we are interested in applying the TP-MOE in optimization and reinforcement learning tasks. For such tasks,  the learning, prediction, and decision making aspects of the model occur in sparse, noisy environments that require heavy-tailed models in order for a learning agent to properly handle the problem at hand. Modelling data with a mixture of Student-$t$ processes is a natural method for dealing with non-stationarity and heavy-tailed errors yet their popularity has still eluded the machine learning community. We seek to fill that gap with the method proposed in this paper. 




\clearpage
\clearpage
\bibliographystyle{apalike}
\bibliography{studentt}
\end{document}